\documentclass{article}
\usepackage[table]{xcolor} 

\usepackage{spconf,amsmath,graphicx}
\usepackage{url}            
\usepackage{booktabs}       
\usepackage{amsfonts}       
\usepackage{nicefrac}       
\usepackage{microtype}      
\usepackage{xcolor}         
\usepackage{graphicx}
\usepackage{multirow}
\usepackage{bbding}
\usepackage{enumitem}
\usepackage[colorlinks,linkcolor=red,anchorcolor=black,citecolor=green]{hyperref}  

\usepackage{cleveref}

\let\OLDthebibliography\thebibliography
\renewcommand\thebibliography[1]{
  \OLDthebibliography{#1}
  \setlength{\parskip}{0pt}
  \setlength{\itemsep}{0pt plus 0.3ex}
}

\begin{document}\sloppy


\def\x{{\mathbf x}}
\def\L{{\cal L}}

\title{A Parallel Attention Network for Cattle Face Recognition}
%

\name{Jiayu Li$^{1, }$$^*$, Xuechao Zou$^{1, }$$^*$, Shiying Wang$^{1}$, Ben Chen$^{1}$, Junliang Xing$^{2}$, Pin Tao$^{1,2, }$$^\dagger$\thanks{$^*$Equal contribution.}\thanks{$^\dagger$Corresponding author.}}
\address{
$^1$Qinghai University, Department of Computer Technology and Applications, Xining, China\\
$^2$Tsinghua University, Department of Computer Science  and Technology, Beijing, China
}


\maketitle

\begin{abstract}
Cattle face recognition holds paramount significance in domains such as animal husbandry and behavioral research. Despite significant progress in confined environments, applying these accomplishments in wild settings remains challenging. Thus, we create the first large-scale cattle face recognition dataset, ICRWE, for wild environments. It encompasses 483 cattle and 9,816 high-resolution image samples. Each sample undergoes annotation for face features, light conditions, and face orientation. Furthermore, we introduce a novel parallel attention network, PANet. Comprising several cascaded Transformer modules, each module incorporates two parallel Position Attention Modules (PAM) and Feature Mapping Modules (FMM). PAM focuses on local and global features at each image position through parallel channel attention, and FMM captures intricate feature patterns through non-linear mappings. Experimental results indicate that PANet achieves a recognition accuracy of 88.03\% on the ICRWE dataset, establishing itself as the current state-of-the-art approach.  The source code is available in the supplementary materials.

\end{abstract}


\section{Introduction}
\label{intro}
With the rapid advancement of computer vision technology, the livestock industry is making strides toward a digital platform wherein artificial intelligence plays a pivotal role in achieving informatization and intelligence in animal husbandry. This drives a swift transformation of traditional farming methods towards greener, more efficient, and more precise.

In the pursuit of achieving meticulous management of cattle, the identification of cattle entities has emerged as a pivotal technological facet. Traditional methods for cattle recognition primarily rely on physical identifiers such as wearing ear tags and embedding chips. However, these conventional methods entail risks of vulnerability, susceptibility to loss, and susceptibility to tampering, potentially causing harm to the animals. Therefore, computer vision and artificial intelligence technologies have been introduced. Non-touch identity recognition methods based on biometric feature images have demonstrated efficient and precise potential in the realm of cattle recognition.

\begin{figure}[t]
    \centering
    \includegraphics[width=1.0\linewidth]{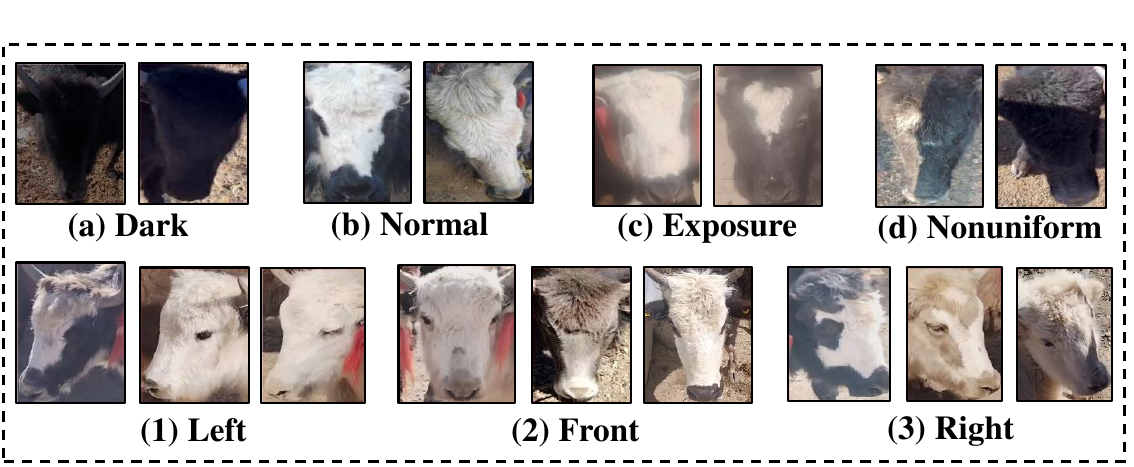}
    \caption{Samples of cattle faces from the dataset exhibiting variations in lighting conditions and different face orientations.}
    \label{fig: data}
\end{figure}

We investigated the primary research methods in the field of cattle recognition. From the retina~\cite{rusk} and iris~\cite{Kong} of the cattle's eyes to the nose print and the overall individual~\cite{bello-dataset, Bhole-dataset}, existing algorithms are predominantly focused on artificially captivity environments, making them challenging to apply in the large-scale wild environment. Due to various environmental factors in the wild, such as different lighting conditions and diverse orientations of cattle faces, the quality of cattle images captured in wild environments is suboptimal, and the background is highly intricate. This renders the direct migration of existing cattle recognition algorithms tailored for confined cattle challenging in the context of the wilderness environment. Furthermore, compared to cattle in artificial captivity, the image data of cattle in real wild scenarios is relatively limited, posing data collection and annotation difficulties.

Hence, we collected and annotated the first large-scale dataset for cattle recognition in a wild environment, ICRWE. Leveraging manually annotated cattle face data, we trained a YOLOv5\footnote{\url{https://github.com/ultralytics/yolov5}}model to obtain a cattle face detection model. Subsequently, we employed this model to crop the cattle faces from the original captured images, thereby diminishing interference from intricate backgrounds and optimizing the dataset images. Our primary contributions can be summarized as follows:

\begin{itemize}
\setlength{\partopsep}{0pt}
\setlength{\topsep}{0pt}
\setlength{\parsep}{0pt}
\setlength{\itemsep}{0pt}
\setlength{\parskip}{0pt}
    \item We propose ICRWE, the first large-scale dataset for cattle face recognition in wild environments. It covers diverse lighting conditions and face orientations. We also annotated and trained a model for cattle face detection to remove the background in the original images.
    \item We introduced a novel parallel attention network, PANet.  The network architecture comprises multiple cascaded Transformer modules with parallel PAM and FMM components. PAM, employing a parallel channel attention mechanism, focuses on the local and global features of the image, while FMM excels in capturing intricate feature patterns through nonlinear mapping. Amalgamating a robust feature from two branches significantly enhanced the performance of cattle recognition.
    \item The experiments demonstrate that PANet achieved a state-of-the-art accuracy of 88.03\% on the challenging real-world wild cattle face recognition dataset (ICRWE).
\end{itemize}

\section{Related Work}

\subsection{Cattle Recognition}
Cattle face recognition is a biometric identification technology that verifies or identifies individuals by analyzing specific biological features of cattle. Currently, methods for cattle recognition primarily encompass biometric approaches such as retinal recognition~\cite{rusk}, iris recognition~\cite{Kong}, nose print recognition, and face recognition~\cite{wang-dataset}. Rusk \emph{et al}. \cite{rusk} through voluntary efforts, conducted manual recognition of cattle and sheep based on retinal images. The results indicated an accuracy rate of 96.2\% in identifying cattle. Kong \emph{et al}. \cite{Kong} proposed the use of iris features for identity recognition. Kumar \emph{et al}. \cite{kumar-deep} utilized texture features to describe the features extracted from different Gaussian pyramid smoothing levels of cattle nose print images. Face recognition methods for cattle involve extracting face image information using deep learning techniques, Xia \emph{et al}. \cite{xia-Cattle} proposed a face description model based on Local binary pattern texture features. They utilized principal component analysis combined with sparse representation classification to recognize the face images of cattle. Zheng \emph{et al}. \cite{Zheng-vit} incorporate a learnable mask matrix into the ViT, utilizing the mask matrix to discern the significance of image blocks.

Existing research in cattle recognition primarily focuses on artificially captivity environments, not meeting the practical demand for large-scale recognition in the wild. Current datasets mainly focus on detection and classification, with a relatively limited number specifically designed for cattle recognition. To address this gap, we created ICRWE, the initial large-scale dataset for cattle face recognition in wild environments.

\subsection{Attention Mechanism}
The attention mechanism plays a crucial role in the field of artificial intelligence, notably in areas such as computer vision~\cite{chen2023leformer,zou2023diffcr,zou2023pmaa,li2020survey}, natural language processing, and speech processing~\cite{li2022efficient,kuo2022inferring}. SENet~\cite{senet} introduced the mechanism of channel attention. By assimilating the significance weights of each channel, it dynamically adjusts the channel weights of the feature map, thereby enhancing meaningful feature representations. In 2018, Woo \emph{et al}.~\cite{cbam} introduced CBAM, combining channel attention and spatial attention mechanisms. CBAM achieves global awareness and organized integration of feature maps by considering the importance of both channel and spatial features. In 2020, Wang \emph{et al}.~\cite{eca} introduced ECA-Net, which presents an efficient channel attention mechanism, improving feature representational capacity by adaptively adjusting channel feature weights. Both ViT~\cite{vit} and Swin~\cite{swin} employ the self-attention mechanism, enabling the model to focus dynamically on information from different positions within the input sequence.
In 2022, Liu \emph{et al}.~\cite{convnet} introduced ConvNeXt, a model that integrates dense connections, group convolutions, and a self-attention mechanism reminiscent of Transformers. This integration captures vital connections between positions, enhancing model stability, generalization, and resistance to interference.
In cattle face recognition, the attention mechanisms are underused. In the wild, low-quality cattle images and complex backgrounds render traditional recognition algorithms impractical. Therefore, we propose a parallel attention network, PANet, for cattle face recognition.

\section{Dataset}
\subsection{Dataset Creation}
\begin{figure}
    \centering
    \includegraphics[width=1\linewidth]{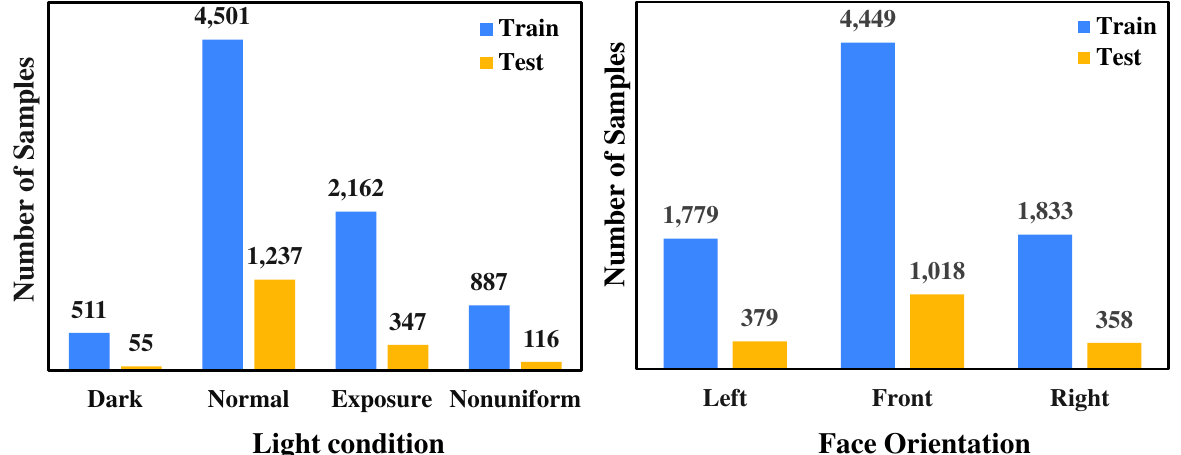}
    \caption{Statistical results for four levels of lighting condition and three face orientations in the training and testing datasets.}
    \label{fig: enter-label}
\end{figure}

We employed a non-intrusive and non-disruptive video recording method for the data collection of cattle behavior. Specifically, we selected cattle in their natural state, freely grazing and in good health, as our study subjects. The data collection period we have spanned from September to October 2023. During the data acquisition process, we maintained a distance of 1 to 3 meters from the cattle, utilizing high-resolution cameras for recording. The filming duration for each individual ranged from 5 to 10 seconds. The collected videos are stored in MP4 format, totaling 483 cattle.
\begin{table}[t]
\centering
\small
\caption{Comparison of our ICRWE dataset with existing datasets for cattle recognition. ``-'' indicates that the original paper did not disclose the corresponding metric.}
\setlength{\tabcolsep}{4.4pt}
\begin{tabular}{c|ccccccc}
\toprule
\textbf{Ref.} & \textbf{IDs} & \textbf{Sample} &\textbf{Size} &
\textbf{P. E.} & \textbf{FDA} & \textbf{LIA}& \textbf{FOA} \\
\midrule
~\cite{bello-dataset} & 136 & 1,237 & 640$\times$480  & Cap. &$\times$&$\times$&$\times$\\
~\cite{Bhole-dataset} & 400 & 4,000 & -        & Cap. &$\times$&$\times$&$\times$\\
~\cite{shen-dataset} & 105 & 1,433 & 640$\times$480  & Cap. & $\checkmark$ &$\times$& $\times$\\
~\cite{wang-dataset} & 36  & 187  & -        & Cap. & $\times$& $\times$&$\times$\\
~\cite{yang-dataset} & 200 & 8,000 & -        & Cap. & $\checkmark$ &$\times$&$\times$\\
~\cite{Zhu-dataset} & 70  & 2,218 & -        & Cap. &$\times$&$\times$ & $\times$\\
\midrule
\rowcolor[RGB]{217,217,217}Ours    & 483 & 9,816 & 720$\times$1,080 & Wild    & $\checkmark$ & $\checkmark$ &$\checkmark$ \\
\bottomrule
\end{tabular}
\label{tab: datasets}
\end{table}
During the video processing stage, we manually curated and removed images with severe motion blur or those without cattle due to camera capture. Subsequently, we annotated the partly images with cattle faces and trained a YOLOv5 to obtain an efficient cattle face detection model. This model performed face detection on the total dataset, cropping the detected cattle faces to create a cattle face recognition dataset. Subsequently, we applied operations like horizontal flipping, brightness adjustment, and translation. Annotations were conducted for lighting conditions and face orientation Fig.~\ref{fig: data}. We categorized the lighting condition into four levels: dark, normal, exposure, and non-uniform. Furthermore, we classified cattle face orientations into front, right, and left. The statistical results are depicted in Fig.~\ref{fig: enter-label}.

\subsection{Dataset Properties}
\begin{figure}
    \centering
    \includegraphics[width=1\linewidth]{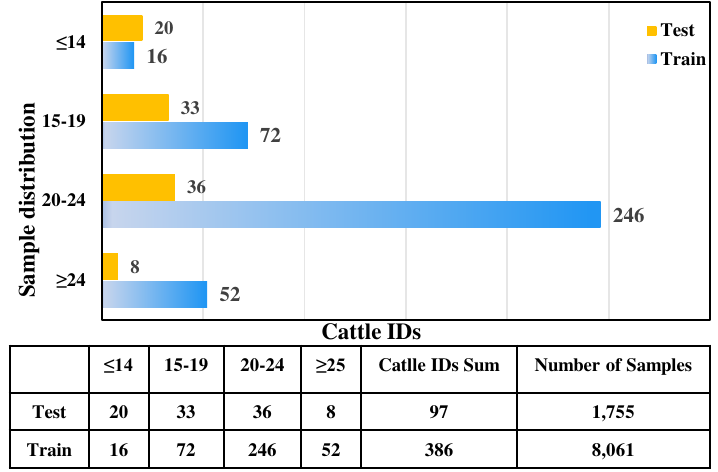}
    \caption{The distribution and statistics of sample quantities for each cattle in both the training and testing sets.}
    \label{fig: sample}
\end{figure}

 The dataset comprises face samples from 483 cattle, totaling 9,816 samples. We address interference factors in wild environments affecting cattle face recognition, considering elements such as lighting conditions and face orientation in dataset annotations. The comparison between our ICRWE dataset and existing datasets for cattle recognition is presented in Table \ref{tab: datasets}. In this chart, P.E. indicates the photography environment, whether captive or in the wild. FDA, LIA, and FOA respectively stand for face annotation, lighting annotation, and face orientation annotation.
As illustrated in Fig.~\ref{fig: sample},  The training set and test set were partitioned in an 8:2 ratio. Additionally, to simulate real-world scenarios, we select one sample from each cattle's test set and include it in the feature library, forming the feature library for the testing set.

\section{Method}
\subsection{Overall Pipeline} \label{sec:pipeline}

\textbf{Problem Definition.} Each cattle possesses its unique identity ID. For the dataset $D=\left \{ \left ( x,y \right ) ,x\in X,y\in Y \right \} $, where $x$ is the cattle face image and $y$ represents the identity ID corresponding to $x$. To simulate real-world scenarios, we establish a feature library $L$, which includes a randomly selected cattle face image  $x_{l}$ and its corresponding identity ID $y_{l}$ for each cattle to be recognized. Each cattle to be recognized has multiple cattle face samples $x_{t}$. By calculating the similarity between $x_{l}$ and $x_{t}$ in the feature repository, we identify the ID $y_{l}$ with the highest similarity to $x_{l}$. Cosine distance is employed to calculate the similarity. If the highest similarity $y_{l}$ matches the ID $y_{t}$ of $x_{t}$ in the feature library $L$, then the recognition is correct. Otherwise, it is considered incorrect. The accuracy is calculated as follows:
\begin{equation}
    ACC=TP/N,
\end{equation}
where $TP$ is the number of correctly identified cattle face samples, and $N$ is the total count of samples to be recognized.
\begin{figure}[t]
    \centering
    \includegraphics[width=1\linewidth]{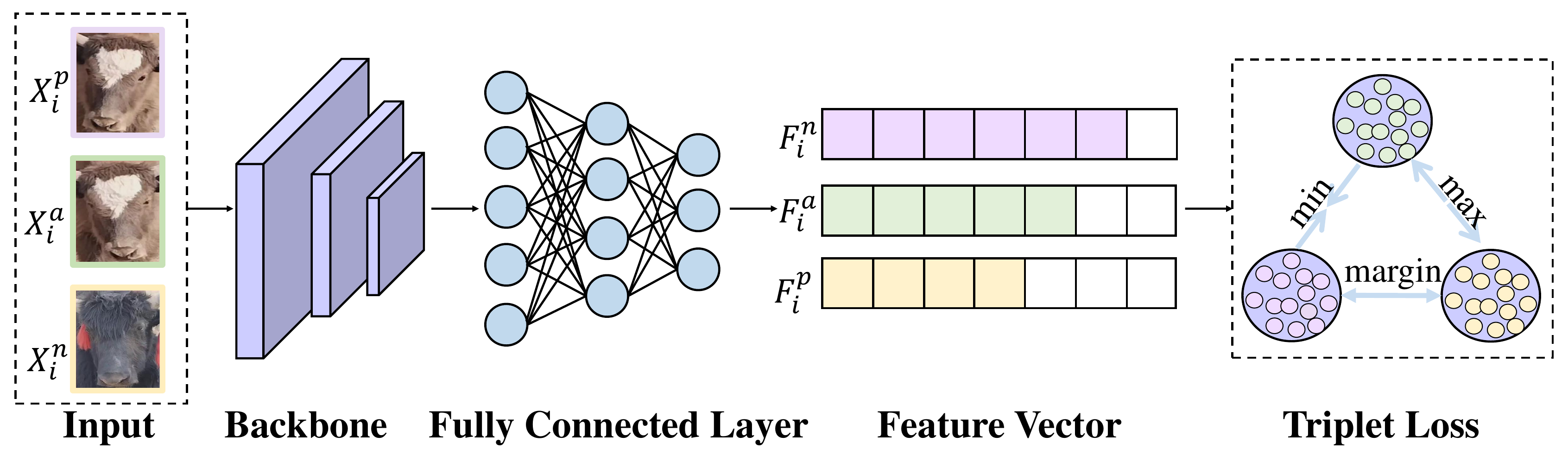}
    \caption{The network framework employed in the method of cattle recognition based on cattle face features.}
    \label{fig: Network}
\end{figure}
\begin{figure*}
    \centering
    \includegraphics[width=1.0\linewidth]{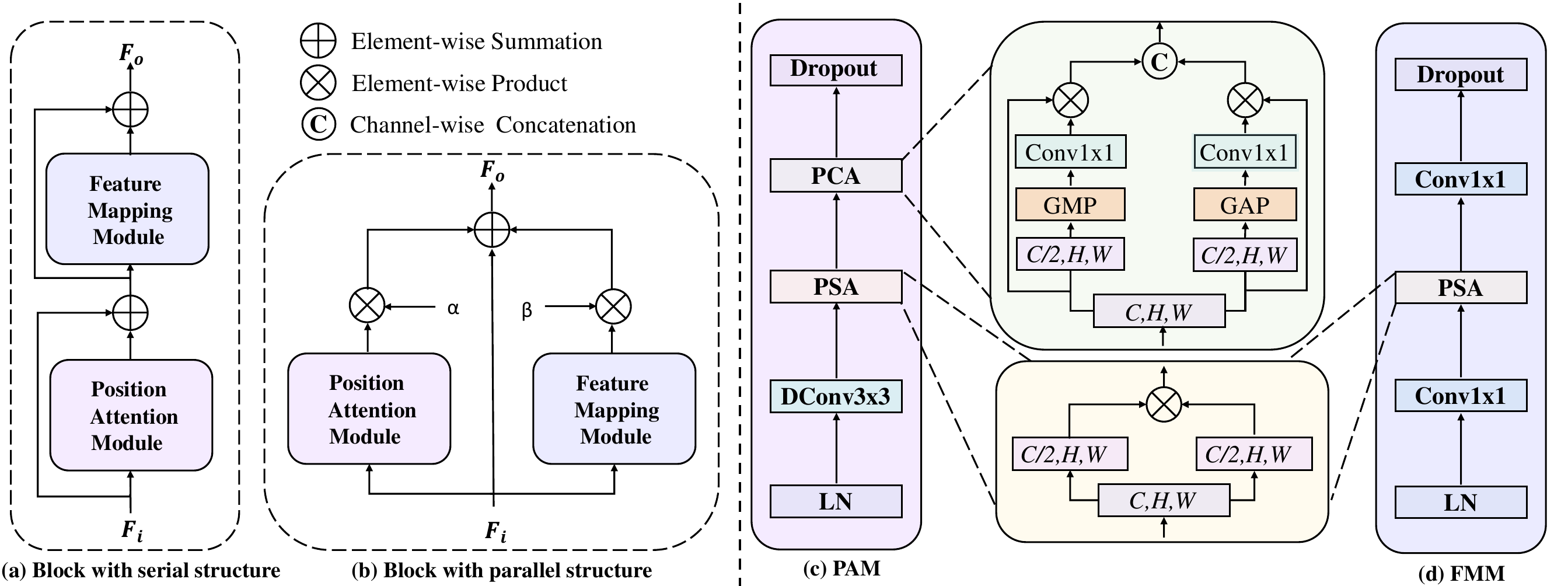}
    \caption{
    (a) Sequential Transformer structure; (b) Parallel Transformer structure; (c) The specific architectural implementation of the Position Attention Modules (PAM); (d) The specific architectural implementation of the Feature Mapping Modules (FMM).}
    \label{fig: Parallel}
\end{figure*}
\\
\textbf{Training Framework.} The overall training framework, as illustrated in Fig.~\ref{fig: Network}, begins by taking the cattle face image as input. The image undergoes feature extraction through the backbone, followed by a fully connected layer that reflects these feature maps into a 128-dimensional vector. Training is performed using a triplet loss function.
 We randomly select a sample $X_{i}^{a}$ from the dataset as an anchor. Simultaneously, we randomly choose samples $X_{i}^{p}$ from the same IDs as the anchor's positive examples and samples $X_{i}^{n}$ from different IDs as negative examples.
We opted for semi-hard~\cite{facenet} triplet selection during training. The formula for the loss function is:
\begin{equation}
    L_{triplet} = \sum_{i=0}^{N-1}\left [ d_{1} -d_{2} + \alpha  \right ] _{+}, 
\end{equation}
where $\alpha $ is a margin that is enforced between positive and negative pairs. $d1$ represents the distance between samples from the same cattle, while $d2$ signifies the distance between samples from different cattle.

\subsection{Proposed Network}
We introduce a novel parallel attention network, PANet, which consists of two integral components: the backbone and the fully connected layer. The backbone is dedicated to cattle face feature extraction, while the fully connected layer maps features into a vector of dimensionality 128.
\begin{table*}[]
\small
\centering
\caption{Accuracy (\%) and efficiency of cattle face recognition on the proposed ICRWE dataset for PANet and existing methods.}
\setlength{\tabcolsep}{7.5pt}
\begin{tabular}{l|cccc|ccc|c|cc}
\toprule
  \multirow{2}{*}{\textbf{Method}} &
  \multicolumn{4}{c|}{\textbf{Light Condition}} &
  \multicolumn{3}{c|}{\textbf{Face Orientation}} &
  \multirow{2}{*}{\textbf{Sum}} &
  \multirow{2}{*}{\textbf{\#M}~$\downarrow$} &
  \multirow{2}{*}{\textbf{\#P}~$\downarrow$}
  \\
  \cmidrule(lr){2-5} \cmidrule(lr){6-8}
  & \textbf{Dark} &\textbf{Normal} & \textbf{Exposure} & \textbf{Nonuiform} & \textbf{ Left} & \textbf{Front} & \textbf{Right} & &  &   \\
\midrule
ResNet50~\cite{resnet}$\textsuperscript{CVPR15}$  &2.10&55.56&15.67&3.13&15.32&47.12&14.07&76.46&4.13&23.76\\
SENet~\cite{senet}$\textsuperscript{CVPR18}$    &2.62&53.33&16.75&5.01&15.67&47.24&14.81&77.72&4.14&26.30\\
MobileNetv2~\cite{mobilenetv2}$\textsuperscript{CVPR18}$  &2.39&55.15&13.45&4.27&13.21&49.80&12.25&75.27&1.17&3.57\\
ShuffleNetv2~\cite{shufflenet}$\textsuperscript{ECCV18}$   &2.79&57.60&10.77&\textbf{5.70}&17.49&49.34&15.72&82.56&0.15&1.38 \\
Res2Net~\cite{res2net}$\textsuperscript{CVPR19}$  &2.16&44.67&12.67&3.64&11.91&38.01&10.60&60.51&4.23&23.27\\
EfficientNet~\cite{efficientnet}$\textsuperscript{PMLR19}$ &2.34&48.66&15.10&4.22&13.73&43.84&12.71&70.31&0.03&0.84\\
HRNet~\cite{hrnet}$\textsuperscript{CVPR19}$&2.16&54.30&15.33&4.67&15.04&47.41&14.02&76.41&4.35&19.51\\
ViT~\cite{vit}$\textsuperscript{ICLR21}$&2.51&52.08&15.84&4.96&15.95&46.38&13.05&75.39&55.40&86.23\\
Swin~\cite{swin}$\textsuperscript{CVPR21}$  &2.22&51.79&15.61&4.78&14.24&46.89&13.27&74.41&4.50&28.95\\
ConvNeXt~\cite{convnet}$\textsuperscript{CVPR22}$  &2.11&57.89&16.46&4.10&14.98&50.54&15.04 &80.57&4.51&29.01\\
\midrule
\rowcolor[RGB]{217,217,217} \textbf{PANet (Ours)} &\textbf{2.80} &\textbf{61.82}&\textbf{18.48} &4.96 &\textbf{18.80}& \textbf{53.10}&\textbf{16.13}& \textbf{88.03} &2.83 &19.61\\
\bottomrule
\end{tabular}
\label{tab:acc}
\end{table*}

\begin{table}[]
\centering
\caption{Ablation study results of different network structures.}
\setlength{\tabcolsep}{9.5pt}
\begin{tabular}{cc|cc|c}
\toprule
\textbf{Serial }&\textbf{ Parallel }&\textbf{ GAP }& \textbf{GMP} &\textbf{ Acc (\%) } \\
\midrule
$\checkmark$ &$\times$ & $\checkmark$&$\times$&   73.11  \\
$\checkmark$& $\times$ &$\times$ &$\checkmark$& 74.02   \\
$\checkmark$ &$\times$ & $\checkmark$&$\checkmark$&   76.35  \\
$\times$& $\checkmark$ &$\checkmark$ &$\times$&   85.52   \\
$\times$& $\checkmark$ &$\times$&$\checkmark$&  86.58     \\ 
\rowcolor[RGB]{217,217,217} $\times$&$\checkmark$ &$\checkmark$& $\checkmark$ &  \textbf{88.03}  \\ 
\bottomrule
\end{tabular}
\label{tab: ablation}
\end{table}

\begin{figure}
    \centering
    \includegraphics[width=1.0\linewidth]{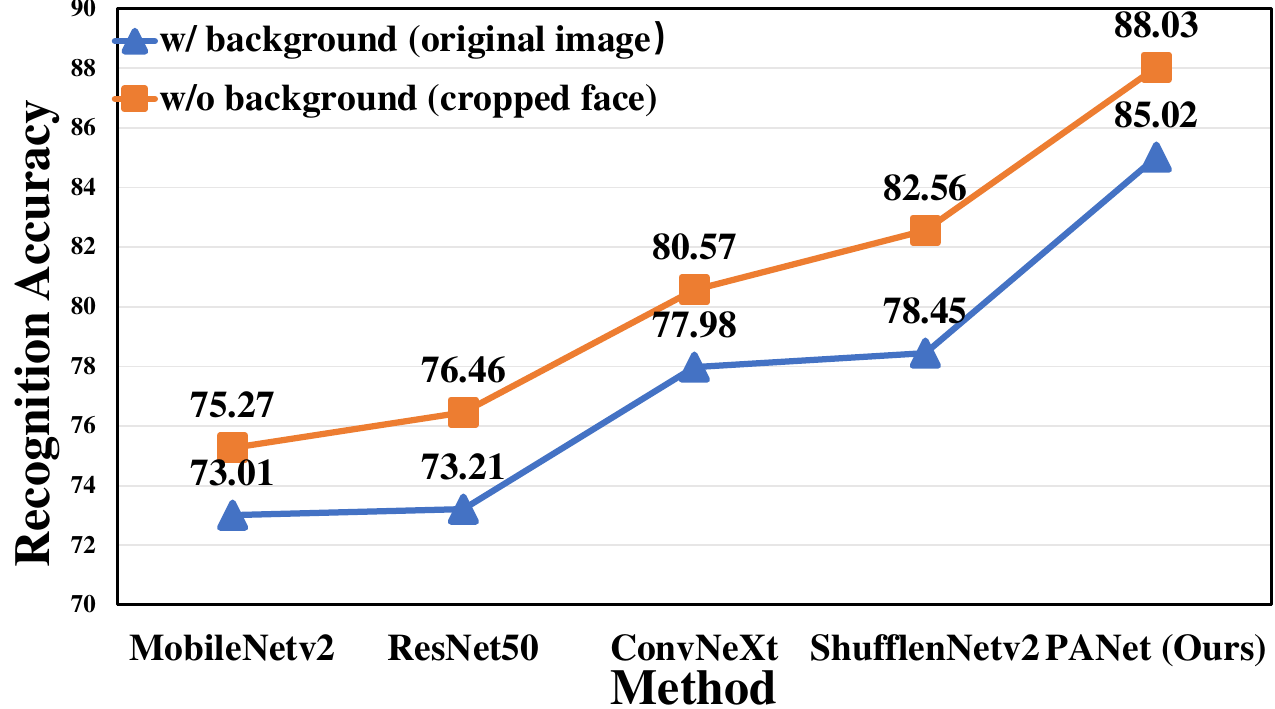}
     \caption{The accuracy of cattle recognition under two input conditions: with background and without background.}
    \label{fig: background}
\end{figure}

Our backbone architecture draws inspiration from the multi-stage design introduced in Swin~\cite{swin}, where each stage comprises several stacked blocks. We have eschewed the conventional serial Transformer architecture depicted in Fig.~\ref{fig: Parallel} (a) and instead introduced a parallel Transformer structure, as illustrated in Fig.~\ref{fig: Parallel} (b). The feature map $\mathbf{F_{i} }\in \mathbb{R} ^{C\times H\times W} $ is simultaneously fed into PAM and FMM for processing, followed by the residual aggregation to yield the output features:
\begin{equation}
  \mathbf{F_o} = \operatorname{PAM}(\mathbf{F_{i} }) \otimes  \alpha  +  \operatorname{FMM}(\mathbf{F_{i} }) \otimes  \beta  +\mathbf{F_{_i} },
\end{equation}
where $\alpha \in \mathbb{R} ^{{C\times 1\times 1} } $ and $\beta \in \mathbb{R} ^{{C\times 1\times 1} } $ are learnable parameters, allowing the model to autonomously learn the importance of each channel to the specific task. $\mathbf{F_{i} }$ represents the input cattle face feature, 
and $\mathbf{F_{0} }$ represents the output feature map.
\\
\textbf{Position Attention Module.} The attention mechanism allows the model to learn the importance of input features at various positions dynamically. We have devised a parallel attention module, such as Fig.~\ref{fig: Parallel} (c). Layer normalization is applied for the input feature $\mathbf{F_{i}}$, followed by feature extraction using depth-wise separable convolution. Subsequently, an approximate gating operation is implemented through the Parallel Self-Activation (PSA)~\cite{NAFNet} mechanism. This operation divides the input tensor into two parts along the channel dimension, followed by element-wise multiplication. Finally, the proposed Parallel Channel Attention (PCA) module is employed for positional attention selection. It divides the input feature map into two equally channel-wise parts, $\mathbf{F_{i}^{1}}\in \mathbb{R}^{\frac{C}{2} \times H\times W}$ and $\mathbf{F_{i}^{2}}\in \mathbb{R}^{\frac{C}{2} \times H\times W}$.
The definition of the PAM is as follows:
\begin{equation}
    \mathbf{F_{pam} }= \operatorname{PCA}\left ( \operatorname{PSA}\left (\operatorname{DConv}_{3\times 3} \left (  \gamma  \left ( \mathbf{F}_{i}  \right )  \right ) \right )  \right ),
\end{equation}
where, $ \gamma$ represents layer normalization, and $\operatorname{DConv}_{3\times3}$ denotes depthwise separable convolution. Some details, such as reshaping and dropout, are omitted for the sake of simplicity. The computation is subsequently performed as follows:
\begin{equation}
    \operatorname{PCA}(\mathbf{F_{i}}) = \varphi   (\mathbf{F_{i}^{1}} \otimes  \operatorname{GMP} \left ( \mathbf{F_{i}^{1}}\right ),\mathbf{ F_{i}^{2} } \otimes \operatorname{GAP}  \left ( \mathbf{F_{i}^{2} } \right )),
\end{equation}
where GMP denotes global max-pooling, preserving locally salient features; GAP represents global average pooling, extracting overall global information; $\varphi$ signifies the concatenation of channel attention results for two distinct pooling methods, amalgamating both global and local information.
\\
\textbf{Feature Mapping Module.} The MLP is pivotal in neural networks; our design of the Multi-Layer Perceptron is illustrated in Fig.~\ref{fig: Parallel} (d). The input image $\mathbf{F_{i}}$ undergoes layer normalization, followed by a convolution operation. Similarly, the gating mechanism of the PSA module is applied, and then a convolution adjusts the feature map dimensions to align with the model's requirements. The definition of FMM is as follows:
\begin{equation}
    \mathbf{F_{fmm}} = \operatorname{MLP}\left ( \gamma \left ( \mathbf{F_{i}}  \right )  \right ),
\end{equation}
where $\operatorname{MLP}$ integrates information across channels, and $\operatorname{PSA}$ functions as the non-linear activation in this context.
\section{Experiments and Results}
This experiment utilized an NVIDIA Tesla V100 with 32GB of RAM, configured with a training cycle of 300 epochs and a batch size of 16. For the stability of the initial learning phase, we set the initial learning rate to 5e-4, with a weight decay of 5e-2. We implemented a cosine annealing learning rate decay strategy to optimize the learning rate. Due to variations in image pixel sizes, we resized the images to $224\times 224$.
\subsection{Comparison with SOTA Methods}
ICRWE takes into account the influence of light conditions and face orientation in cattle face recognition. The experimental results are presented in Table \ref{tab:acc},  \#M (MACs, G) and \#P (Params, M) denote the number of multiply-accumulate operations and the number of parameters, respectively. Regarding lighting conditions, PANet consistently outperforms other methods across various lighting conditions, particularly excelling in normal scenarios with improvements of 3.93\%, respectively, compared to suboptimal methods. Regarding face orientation, PANet notably enhances recognition accuracy by 1.31\% and 2.56\% in left and front orientations, respectively, compared to the ConvNeXt. PANet demonstrates outstanding performance, achieving an overall recognition accuracy of 88.03\%. Furthermore, PANet exhibits a relatively lower parameter count and computational complexity. These results thoroughly validate the rationale behind our dataset design and underscore the potential value of PANet in practical applications. 

\subsection{Ablation and Analysis}
We conducted ablation experiments to investigate the impact of connectivity and attention modules on the performance of our proposed model for cattle recognition. The results are as depicted in Table \ref{tab: ablation}. We observed a significant performance improvement of 11.68\% under the Parallel connectivity scheme. Simultaneously employing GAP and GMP attention modules represents an improvement of 2.51\% over using GAP alone and 1.45\% over using GMP alone.
We employed various models to experimentally compare images with the original background retained and images with removed backgrounds. According to the experimental results depicted in Fig.~\ref{fig: background}, we observed a significant improvement in cattle face recognition accuracy after all methods' background cropping. The empirical evidence substantiates the efficacy of background cropping in enhancing the dataset during its curation process.
\section{Conclusion}
This paper introduces the first large-scale cattle recognition dataset, ICRWE, designed explicitly for wild environments, alongside a parallel attention network, PANet. These contributions enrich the data resources and algorithmic approaches in the field of cattle recognition, and will have a positive impact on practical livestock applications in wild environments.
\section{Acknowledgements}
This work was partly supported by the National Natural Science Foundation of China under Grant No. 62222606 and 62076238.
\bibliographystyle{IEEEbib}
\bibliography{icme2023template}
\end{document}